\pdfoutput=1

\documentclass[11pt]{article}

\usepackage{EMNLP2023}

\usepackage{times}
\usepackage{latexsym}
\usepackage{lmodern}

\usepackage[T1]{fontenc}

\usepackage[utf8]{inputenc}

\usepackage{microtype}

\usepackage{inconsolata}
\usepackage{xcolor}
\usepackage{enumitem}
\usepackage{booktabs}
\usepackage{array}
\usepackage{hyperref}
\usepackage{url}
\usepackage{multirow}
\usepackage{colortbl}
\usepackage{booktabs}
\usepackage{amssymb}
\usepackage{amsmath}
\usepackage{amsthm}
\usepackage{graphicx}
\usepackage{makecell}
\usepackage{threeparttable}
\usepackage{CJKutf8}
\usepackage{lscape}
\usepackage{fancyhdr}
\usepackage[ruled,vlined]{algorithm2e}

\usepackage{threeparttable}
\usepackage{wrapfig}
\usepackage{subfigure}
\usepackage{EMNLP2023}
\usepackage{algorithmic}
\usepackage{CJKutf8}
\usepackage{mdframed}
\usepackage{arydshln}

\title{RIVAL: Reinforcement Learning with Iterative and Adversarial Optimization for Machine Translation}

\author{
    {\normalsize
    \ \ \textbf{Tianjiao Li}$^{\bigstar}$,
     \textbf{Mengran Yu}$^{\bigstar}$, 
     \ \ Chenyu Shi$^{\bigstar\blacklozenge}$, 
     }\\
     {\normalsize
     \ \ \textbf{Yanjun Zhao}$^{\diamondsuit}$\thanks{This work was done when Yanjun Zhao and Chenyu Shi was an intern at Bilibili Inc.},
     \ \ \textbf{Xiaojing Liu}$^{\bigstar}$,
     \ \ \textbf{Qiang Zhang}$^{\bigstar}$\thanks{$^\dagger$ Corresponding Author}
     }\\
     {\normalsize
     \textbf{Qi Zhang}$^{\blacklozenge}$,
     \ \ \textbf{Xuanjing Huang}$^{\blacklozenge}$,
     \ \ \textbf{Jiayin Wang}$^{\bigstar}$
     }\\
    {$^\bigstar$Bilibili Inc., Shanghai, China} \\
    {$^\diamondsuit$Xi'an Jiaotong University, China} \\
    {$^\blacklozenge$School of Computer Science, Fudan University, China} \\
    \texttt{mengranyu97@gmail.com, chenyushi22@m.fudan.edu.cn} \\
}

\begin{document}
\maketitle
\begin{abstract}
Large language models (LLMs) possess strong multilingual capabilities, and combining Reinforcement Learning from Human Feedback (RLHF) with translation tasks has shown great potential.
However, we observe that this paradigm performs unexpectedly poorly when applied to colloquial subtitle translation tasks.
In this work, we investigate this issue and find that the offline reward model (RM) gradually diverges from the online LLM due to distributional shift, ultimately leading to undesirable training outcomes.
To address this, we propose RIVAL, an adversarial training framework that formulates the process as a min–max game between the RM and the LLM. RIVAL iteratively updates the both models, with the RM trained to distinguish strong from weak translations (qualitative preference reward), and the LLM trained to enhance its translation for closing this gap. To stabilize training and improve generalizability, we also incorporate quantitative preference reward (e.g., BLEU) into the RM, enabling reference-free quality modeling aligned with human evaluation.
Through extensive experiments, we demonstrate that the proposed adversarial training framework significantly improves upon translation baselines.
\end{abstract}

\section{Introduction}
\label{sec_introduction}

Recent advances in pre‑trained large language models (LLMs) have yielded state‑of‑the‑art results across a wide range of benchmarks~\cite{achiam2023gpt, grattafiori2024llama, bai2023qwen}.
In the field of neural machine translation (NMT), researchers have also been exploring ways to leverage the powerful capabilities of LLMs to improve translation quality.
Although most existing methods\cite{wiseman-rush-2016-sequence, ranzato2015sequence} rely on Maximum Likelihood Estimation (MLE)-based supervised fine-tuning, they suffer from exposure bias, leading to error accumulation and degraded translation quality. Moreover, such models often lack global coherence due to the limited modeling of sentence-level context\cite{kiegeland-kreutzer-2021-revisiting, kreutzer-etal-2018-neural}. These issues have catalyzed growing interest in alternative approaches such as Reinforcement Learning from Human Feedback (RLHF)\cite{he2025r1t1fullyincentivizingtranslation,feng2025mtr1zeroadvancingllmbasedmachine,tan2025remedylearningmachinetranslation}.

\begin{figure}[t]
\centering
  \includegraphics[width=3in]{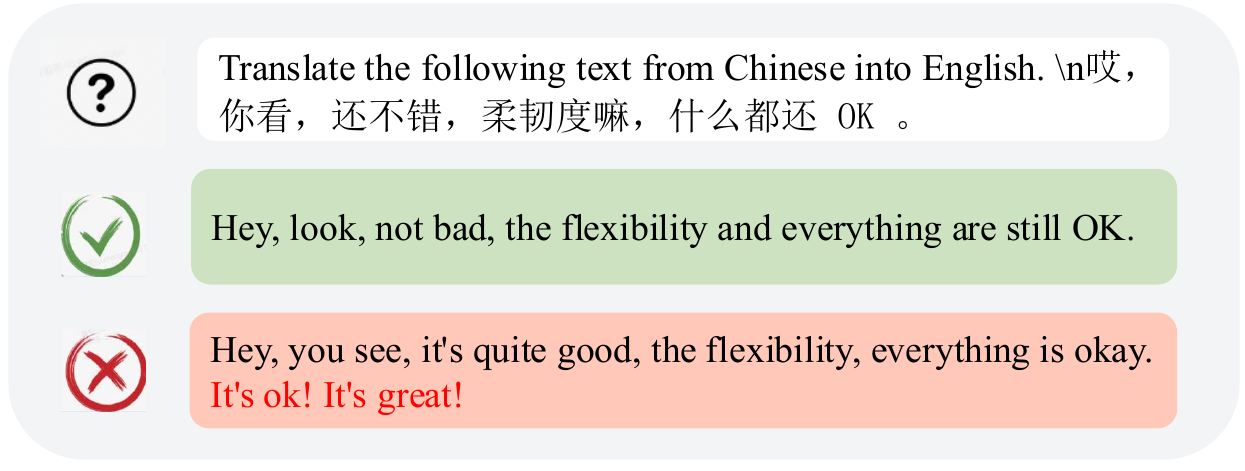}
  \caption{The illustration highlights the issue arising when directly applying vanilla RLHF to a colloquial subtitle dataset. The weak translator tends to generate content that is not present in the source text but receives a favorable score from the RM.}
 \label{first_pic}
 \vspace{-2mm}
\end{figure}

Currently, most NMT systems\cite{mohiuddin2022data, bansal2022data} focus on formal and written language, with limited research addressing loosely structured and colloquial subtitles. To bridge this gap, we first construct a dedicated dataset and apply RLHF to optimize the translation quality for this task. This dataset is characterized by its diversity in both linguistic styles and domains.
However, we find that the performance of vanilla RLHF is suboptimal. As shown in Figure \ref{first_pic}, the model tends to generate content that does not exist in the source text for cheating the RM to obtain a high score, which is usually called reward hacking.
Through analysis, we identify that this issue arises from the continuous distributional shift\cite{touvron2023llama, luo2024sambo} of the LLM during RL training, indicating that offline RM is increasingly ineffective in providing reliable supervision signals, particularly in the context of online RL training.

To address the issue of distribution shift, we propose RIVAL, (\textbf{R}einforcement learning with \textbf{I}terative and ad\textbf{V}ersari\textbf{AL} optimization) an approach inspired by the adversarial training paradigm of generative adversarial networks (GANs)~\cite{goodfellow2014generative}. 
We reformulate the conventional RLHF training scheme as an explicit min–max game between the RM and the LLM. 
The RM is trained to maximize the score gap between the translation pairs produced by strong translators and weak translators. We name this signal the qualitative preference reward.
The LLM training objective is to minimize the gap in translation quality between itself and strong translators.
By iteratively optimizing both models, the RM is continuously updated to adapt to the distributional shifts introduced during LLM training.

In experiments, we further observe that as training progresses, the RM may learn a distribution divergent from the true reward signal. We attribute this issue to the large exploration space inherent in translation tasks, where purely qualitative preference reward may not consistently ensure beneficial optimization. To mitigate this, we introduce quantitative preference rewards (e.g., BLEU\cite{papineni2002bleu}) into our adversarial framework to stabilize the iterative training process.
However, directly using these signals suffers from poor generalizability\cite{sellam2020bleurt} and makes them unsuitable as dynamic rewards. We extend the RM to simultaneously predict this type of quantitative preference reward for incorporating them into our adversarial framework.
This approach enables more generalized reference-free modeling of translation quality while maintaining alignment with human evaluation criteria through indirect reference guidance.

Our core contributions are as follows:
\vspace{-3mm}
\begin{enumerate}
\item In this work, we propose an iterative and adversarial RL approach named RIVAL, which alternately optimizes the RM and the LLM through a competitive process, progressively improving performance from weak to strong by incorporating both qualitative and quantitative preference rewards.
 
\item Through extensive experiments, we demonstrate that our RIVAL effectively improves in-domain translation quality without compromising the out-of-distribution performance of the model.
 
\item We provide a simple and effective pipeline for processing colloquial Chinese-English parallel subtitle translation datasets and release the processed data as an open-source resource.
\end{enumerate}

\section{Background}
\label{sec_Background}

In the Bradley–Terry model~\citep{TB_model}, a pairwise choice between items $i$ and $j$ depends on their latent utilities. The probability of selecting the item $i$ increases monotonically with its utility relative to that of $j$. Formally, these probabilities are obtained by applying a softmax function to the log‑utilities $r(\cdot)$ of the items.

\begin{align}
\label{eqn:BT}
    Q(i\succ j) &= \frac{\exp(r(i))}{\exp(r(i))+\exp(r(j))}  \\ \notag
 &= \text{softmax}(r(i),r(j)).
\end{align}

While this formulation elegantly captures static preference judgments, real-world applications—particularly in natural-language generation often require aligning model behavior with nuanced human values. RLHF addresses this gap by leveraging human evaluators to provide comparative or scalar feedback on model outputs. First, a RM is trained to predict these human judgments from pairs or ratings of model candidates \citep{RLHF,stiennon2020learning}. Next, the LLM is fine-tuned via policy optimization methods such as Proximal Policy Optimization (PPO) to maximize the learned reward signal \citep{schulman2017proximal}. This two-stage process has proven effective in producing more helpful, truthful, and harmless generations in large-scale language systems \citep{ouyangRLHF}. In this paper, RLHF is not exclusively limited to human feedback, but also encompasses model-generated feedback.


\section{Pilot Experiment and Analysis}
\label{sec_Pilot_Experiment_Analysis}

In this section, we first explore the application of vanilla RLHF to NMT and conduct a preliminary analysis of the experimental results.
\subsection{How to Apply RLHF to NMT?}

Inspired by RLHF, we model the translation quality problem as a rank-wise comparison between a weak translator and a strong translator. 
Here, we refer to the translation model being trained as a weak translator, while grouping together gold references and outputs from stronger models under the term strong translator.
Accordingly, we adopt the RM framework to train an evaluator for translation quality $r_\phi$, as formulated below:
\begin{align}
\mathcal{L}_{\text{rank}}(r_\phi; \mathcal{D}_{\mathrm{RM}}) = -\mathbb{E}_{\mathcal{D}_{\mathrm{RM}}} \left[ \log Q(y^s \succ y^w \mid x) \right],
\end{align}
where $\mathcal{D}_{\mathrm{RM}}={(x, y^s, y^w)}$ is parallel translation dataset, $y^s$ and $y^w$ are the strong translation and the weak translation of the text $x$, respectively. 

With a learned RM, we first sample a set of candidate responses $\{y_1, y_2, \ldots, y_G\}$ from the previous policy $\pi_{\theta_{\text{old}}}$ when presented with an original text $x$. Then the advantage function $A_i$ is calculated by normalizing each individual reward relative to the distribution of all rewards in the group. Specifically: $A_i = \frac{r_i - \mu_r}{\sigma_r}$
where $\mu_r = \text{mean}(\{r_1, r_2, \ldots, r_G\})$ represents the average of rewards, $\sigma_r = \text{std}(\{r_1, r_2, \ldots, r_G\})$ denotes the standard deviation, and $\{r_1, r_2, \ldots, r_G\}$ are the rewards scored by the learned RM. Finally, GRPO\cite{shao2024deepseekmath} aims to maximize the following objective function to optimize $\pi_{\theta}$:  
\begin{equation}
\begin{aligned}
J_{\mathrm{GRPO}}(\theta) 
&= \mathbb{E}_{x \sim P(X),\, \{y_i\}_{i=1}^G \sim \pi_{\theta_{\mathrm{old}}}(Y \mid x)} \\
&\Biggl[
  \frac{1}{G} \sum_{i=1}^G
  \min\!\Bigl(
    \frac{\pi_{\theta}(y_i \mid x)}{\pi_{\theta_{\mathrm{old}}}(y_i \mid x)}\,A_i,\, \\
    &\mathrm{clip}\!\Bigl(
      \frac{\pi_{\theta}(y_i \mid x)}{\pi_{\theta_{\mathrm{old}}}(y_i \mid x)},
      1-\varepsilon,\,
      1+\varepsilon
    \Bigr)
    A_i
  \Bigr)  \\
  &-\,\beta\,D_{\mathrm{KL}}\bigl(\pi_{\theta}\,\big\|\,\pi_{\mathrm{ref}}\bigr)
\Biggr],
\end{aligned}
\label{eq: GRPO}
\end{equation}
where the hyperparameter $\varepsilon$ controls the PPO clipping threshold and $\beta$ penalizes the Kullback–Leibler (KL) divergence between the optimized policy $\pi_{\theta}$ and the initial policy $\pi_{ref}$. 

According to the approach mentioned above, we conduct pilot experiments applying RLHF to NMT in the task of colloquial subtitle translation. We detail the construction of a diverse and colloquial subtitle translation dataset drawn from real-world videos. The data processing pipeline follows a systematic approach:

1. Subtitle Extraction. We collect real-world videos and employ Automatic Speech Recognition technology to transcribe spoken content. The transcriptions are then segmented into discrete sentences based on natural pauses and semantic completeness. 

2. Subtitle Processing. We remove background music and semantically insignificant utterances. To enhance contextual coherence and conversational continuity, we structure the data into groups of ten sentences. For clarity and standardization, our prompt instructs models to follow in JSON format.

3. Translation Generation. Given the absence of golden translations for real-world video subtitles, we implement a dual-model approach to generate translation pairs of varying quality. We utilize GPT-4o to produce strong translations that served as targets and employ Qwen2.5-7B-Chat as our baseline model to generate weak translations, representing the initial performance that we aim to improve. 

4. Dataset Construction. We filter out low-quality translation outputs and eliminate samples where strong and weak translations exhibite excessive similarity. The filtered prompts and paired translations are incorporated into the RL corpus, while the combinations of prompts and strong translations form the supervised fine-tuning (SFT) dataset.

We provide several cases in Appendix \ref{sec_subtitle_appendix}. Based on the datasets above, we conduct experiments on optimizing NMT using RLHF. Since colloquial translation prioritize free translation and semantic alignment rather than requiring word-for-word correspondence with the target text, we employ COMETkiwi\cite{rei2022cometkiwi} and model judgment, such as GPT-4o, as evaluation metrics. 

\subsection{Why Vanilla RLHF Fails in Colloquial Subtitle Translation?}

Upon examination of the results, the RM achieves a high accuracy rate of $99.5\%$, attributed to the homogeneity of the data distribution. Subsequently, we apply this RM to GRPO training. Despite continuous improvement in the reward curve and no significant abnormalities in translation length, the evaluation results fall short compared to models trained via SFT using identical source texts. We even observe reward hacking phenomena where high reward scores correspond with low translation quality. For example, the model translates non-existing original texts to cheat the RM into assigning high scores. 

\begin{figure}[ht]
\centering
  \includegraphics[width=3in]{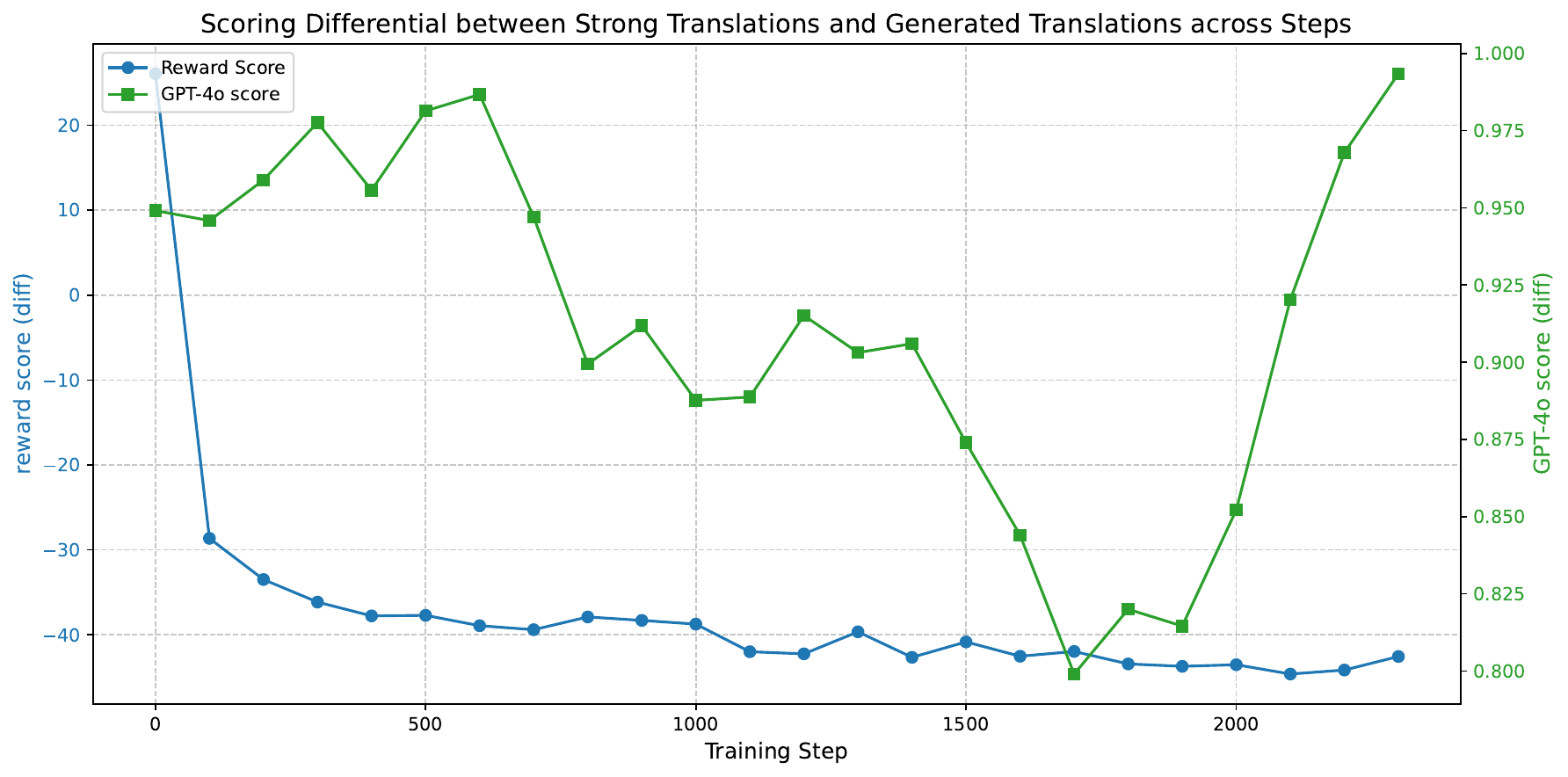}
  \caption{Scoring Differential of the RM and GPT-4o between strong translations and the translations generated by the optimized model.}
 \label{SD}
 \vspace{-2mm}
\end{figure}

To investigate the causes of the aforementioned phenomena, Figure \ref{SD} shows the scoring differential between strong translations and the translations generated by the optimized model, which are evaluated by the RM and GPT-4o. As training progresses, the scoring differential from the RM gradually increases, suggesting that the RM perceives a progressive improvement in the model's generated translations. However, the scoring differential from GPT-4o exhibits an initial decrease followed by a subsequent increase. The significant discrepancy between these two evaluation metrics indicates that the RM trained solely on weak translations from the initial model struggles to adapt to the distributional shift occurring in the training process.

\section{Methods}
\label{sec_methods}
Based on the experimental findings in the previous section, we observe that vanilla RLHF is not well-suited for colloquial subtitle translation tasks. Therefore, inspired by the principles of GANs\cite{goodfellow2014generative}, we propose an adversarial RL approach, RIVAL, that enables both the RM and the LLM to progressively improve from weak to strong. 

\begin{figure*}[ht]
\centering
\includegraphics[width=1\linewidth]{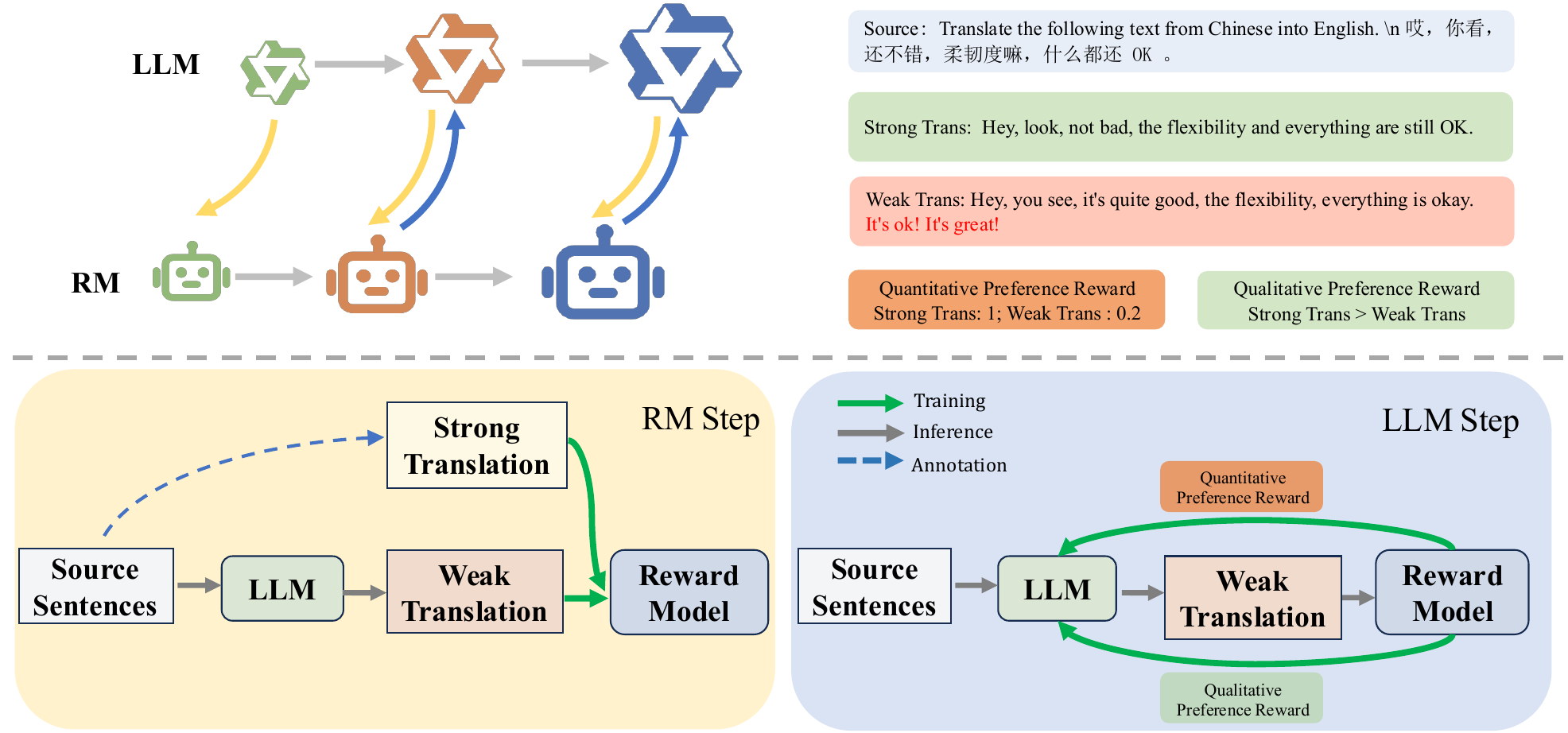}
\vspace{-7mm}
\caption{The RIVAL framework. In the RM updating step, the RM is trained to maximize the score gap between translation pairs produced by strong and weak translators. In the LLM updating step, the LLM is to alleviate the quality gap between its own outputs and those of the strong translators.}
\label{fig:intro}
\end{figure*}

Specifically, we formulate the two-stage training process of RLHF as a min-max adversarial game between LLM $\pi_\theta$ (Generator) and RM $r_\phi$ (Discriminator) , as expressed by the following objective:
 
\begin{align}
    \min_{r_\phi} \max_{\pi_\theta} \ &\mathbb{E} [r_\phi(x, \pi_\theta(y|x))]  \nonumber  
    -  \mathbb{E}_{y \in P_\text{strong}} [r_\phi(x, y)] \nonumber \\
  \textit{s.t. \ } & \text{KL}[\pi_\theta(y|x) \Vert \pi_\text{ref}(y|x)] < \eta, \label{eq:min-max-preference-optim} 
\end{align}
where $P_{\text{strong}}$ denotes the output distribution of a strong translator.
The LLM, as a weak translator, is trained to minimize the gap between its output distribution and that of the strong translator. In contrast, the RM is trained to maximize this distributional gap by distinguishing between the outputs of the weak and strong translators.

By iteratively optimizing both models and using the current LLM to reconstruct new training data for the RM, the RM can effectively learn to serve as a high-quality evaluator for translation quality. Meanwhile, a well-trained RM can in turn provide effective guidance for the LLM, enabling it to explore the open-ended output space and progressively learn to become a strong translator.

\subsection{RM and LLM Optimization Step}
In this subsection, we provide a detailed explanation of the RM and LLM optimization steps.
\subsubsection{RM Step}
The goal of the RM step is to train a high-quality RM to accurately evaluate the quality of the translation. In this step, we keep $\pi_\theta$ fixed and optimize only $r_\phi$.
Therefore, the objective \ref{eq:min-max-preference-optim} simplifies to the following form:
\begin{align}
    \min_{r_\phi} &\mathbb{E} [r_\phi(x, \pi_\theta(y|x))] -  \mathbb{E}_{y \in P_\text{strong}} [r_\phi(x, y)].\label{eq:Discriminator-Step} 
\end{align}

This objective means that the model aims to maximize the gap between the strong translator and the current-round translator.
In the context of our translation task, we first collect source texts $x$ and their corresponding outputs $y$ from a strong translator—either a more powerful model or human experts—from dataset $\mathcal{D}_{RM}$. We then use the current-round LLM $\pi_\theta$ to translate these source texts, obtaining the weak translator’s outputs $\pi_\theta(y|x)$. At this stage, we apply a simple filtering process to exclude samples where the translations from the weak and strong translators are already highly similar, as they offer limited learning signal.
Therefore, we can derive Equation \ref{eq:Discriminator-Step} as follows:
\begin{align}
   \min_{r_\phi} \ &\mathbb{E} [r_\phi(x, \pi_\theta(y|x))] -  \mathbb{E}_{y \in P_\text{strong}} [r_\phi(x, y)] \notag \\
=  \min_{r_\phi} \ &\mathbb{E} [r_\phi(x, y_{weak})] -  \mathbb{E} [r_\phi(x, y_{strong})]  \notag \\
=  \max_{r_\phi} \ &\mathbb{E} [r_\phi(x, y_{strong}) - r_\phi(x, y_{weak})].
\label{eq:Discriminator-Step-Final}
\end{align}

The objective in Equation \ref{eq:Discriminator-Step-Final} aligns with the structure of the RM, and thus we directly adopt the rank loss as the training loss for the RM step. 
In this work, we refer to the rank loss as a qualitative preference loss.
Furthermore, during subsequent iterations, we not only utilize translations generated by the current-round LLM but also replay a subset of outputs from previous rounds. This approach helps prevent excessive distributional shifts, enhances data diversity, and effectively improves the model's robustness.

\begin{algorithm}[ht]
    \caption{RIVAL} 
    \label{alg: RL} 
    \begin{algorithmic}[1]
        \REQUIRE $\mathcal{D_{RM}}$, $\mathcal{D^*_{RM}}$,  $\mathcal{D_{LLM}}$, $\pi_{\theta}$, $r_{\phi}$, iterations $N$, Training steps $T_{RM}, T_{LLM}$ for RM and LLM 
        \FOR{$k=1,...,N$}
            \STATE \CommentSty{//Process Data and Get BLEU.}
            \FOR{$i=1,...,|\mathcal{D_{RM}}|$}
                \STATE Sample piece $(x, y^s_i, y^w_i) \subset \mathcal{D_{RM}}$
                \IF{$sim(y^s_i, y^w_i) < \tau$}
                \STATE \CommentSty{//Get BLEU for translator.}
                \STATE Update $\mathcal{D^*_{RM}} \leftarrow (x, BLEU(y^s_i, y^s_i), BLEU(y^s_i, y^w_i))$
                \ENDIF
            \ENDFOR
            
        \STATE \CommentSty{//RM Step.}
        \FOR{$t=1,...,T_{RM}$}
            \STATE Sample batch $\mathcal{B_{RM}} \subset \mathcal{D^*_{RM}}$\\
            \CommentSty{//Update RM using Equation \ref{eq:multi_head_rankloss}}
            \STATE $\phi_t=\phi_{t-1}-\eta_t^{RM}\ast\nabla_{\phi}\mathcal{L}_{\mathrm{RM}}(\phi_{t-1})$
        \ENDFOR

        \STATE \CommentSty{//LLM Step.}
        \FOR{$t=1,...,T_{LLM}$}
            \STATE Sample $\mathcal{B_{LLM}} \subset \mathcal{D_{LLM}}$
            \STATE Generate $\hat y = LLM_{\theta_t}(x), x \in \mathcal{B_{LLM}}$
            \STATE Compute rewards $r = \mathrm{RM}_{\phi_k}(x, \hat y)$
            \CommentSty{//Update LLM using Equation \ref{eq: GRPO}}
            \STATE $\theta_t = \theta_{t-1} - \eta_t^{LLM}\ast\nabla_{\theta}J(\theta_{t-1})$ \\
            
        \ENDFOR
        \STATE \CommentSty{//Process new RM Training Data.}
        \STATE Update $\mathcal{D_{RM}} \leftarrow (x, y^s,LLM_{\theta_k}(x))$
    \ENDFOR 
    \STATE \textbf{return} $\theta_{LLM}, \phi_{RM}$
    \end{algorithmic} 
\end{algorithm}

\subsubsection{LLM Step}
In the LLM update step, we similarly keep the reward model $r_\phi$ fixed and update only the parameters of the LLM $\pi_\theta$:
\begin{align}
    \max_{\pi_\theta} \ &\mathbb{E}_{x \in D_{LLM}} [r_\phi(x, \pi_\theta(y|x))]  \nonumber  \\
  \textit{s.t. \ } & \text{KL}[\pi_\theta(y|x) \Vert \pi_\text{ref}(y|x)] < \eta. \label{eq:Generator-Step} 
\end{align}

The objective of this training step is to optimize the LLM using the signals provided by the RM. This procedure is aligned with the standard preference learning paradigm, where the LLM is guided to produce the outputs preferred by the RM. In line with the earlier stages of our pipeline, we continue to adopt the GRPO algorithm as our RL method in this phase.

\subsection{Incorporate Quantitative Preference Reward}
Moreover, our experiments reveal that using only the qualitative preference reward leads to instability across training iterations. This may be attributed to the large exploration space of the translation model, as discussed in detail later in the paper.
To address this, we introduce a quantitative preference reward to stabilize and align the optimization trajectory of the model throughout the training iterations. Given that the qualitative preference reward emphasizes semantic alignment, we opt for BLEU as a quantitative preference reward to facilitate lexical alignment. Specifically, rather than directly employing BLEU as a reward signal, we train a RM to approximate BLEU scores.

The key insight is that reference translations may contain errors \cite{xu2024contrastive} or lack diversity, resulting in biased BLEU calculations that lack robustness. Prior work\cite{benedetti2024training, de2022make, bishop1995training} show that noisy-based training can, to some extent, be equivalent to Tikhonov regularization, effectively mitigating overfitting and improving model robustness. Therefore, this work employs noisy data to train a RM that approximates BLEU scores, resulting in reduced sensitivity to noise and enhanced robustness.

Moreover, instead of training a separate model for quantitative preference reward, we leverage the same model backbone used for qualitative preference reward and introduce an additional output head dedicated to quantitative preference reward prediction.
This design offers two key benefits: it reduces the computational cost of model training and enables mutual learning between the two tasks with different supervision forms, thereby mitigating the risk of over-optimization\cite{ahmed2024scalable}.
Finally, our multi-head RM loss function incorporating BLEU is defined as follows:
\begin{equation}
\begin{aligned}
\mathcal{L}_{RM}  &= \mathcal{L}_{\text{qualitative}} +  \alpha \ \mathcal{L}_{\text{quantitative}} \\ 
&= \mathcal{L}_{\text{rank}} + \alpha \ \mathcal{L}_{\text{MAE}}(r_\phi; \mathcal{D}_{\mathrm{RM}}) \\ 
& = \mathcal{L}_{\text{rank}} + \alpha \ \mathbb{E}_{\mathcal{D}_{\mathrm{RM}}} \left[ \left| y^{s}_{\text{BLEU}} - y^{w}_{\text{BLEU}} \right|\right], \label{eq:multi_head_rankloss} 
\end{aligned}
\end{equation}
where $\alpha$ is a coefficient used to balance the weights of the two components. In this paper, we do not assign it a specific value and thus set it to 1. 
Additionally, we recommend using Mean Absolute Error (MAE) as the loss function instead of Mean Squared Error (MSE); this choice will be further discussed in experiments.
The detailed algorithmic procedure can be found in Algorithm \ref{alg: RL}.


\section{Experiment}
\label{sec_experiment}
We will introduce experimental setup, main results and analysis in this section.
\definecolor{gray}{HTML}{E7EAEF}
\definecolor{blue}{RGB}{230,243,254}
\definecolor{red}{RGB}{254,244,181}
\definecolor{green}{RGB}{0,255,255}
\colorlet{gray}{gray!50!white}
\colorlet{red}{red!40!white}
\colorlet{blue}{blue!30!white}
\colorlet{green}{green!10!white}

\subsection{Experimental Setup}
In this subsection, we present the experimental setup and implementation details.

\noindent \textbf{Datasets.} 
Our main experiments are conducted on the WMT dataset and our proposed subtitle dataset. 
For the WMT dataset, we collect Chinese(ZH)-English(EN) parallel corpora from WMT 2017-2020 and perform basic preprocessing, removing sentence pairs with fewer than 30 characters\cite{xu2024a,feng-etal-2024-ladder}.
Additionally, we use English(EN)-German(DE) and Chinese(ZH)-German(DE) medical translation tasks from the WMT dataset as out-of-distribution (OOD) language settings for evaluation.

\noindent \textbf{Evaluation Metrics.}
On the subtitle dataset, we use COMETKiwi and GPT-4o for evaluation because (1) subtitle tasks prioritize semantic content, making BLEU less appropriate; and (2) the references are GPT-4o-generated, making it infeasible to compute BLEU scores against genuine reference translations. For the WMT dataset, both BLEU and COMETKiwi are used to capture lexical fidelity and semantic adequacy.

GPT-4o-based scoring follows the LLM-as-a-Judge\cite{lee2024fleur, weng2022large} paradigm, evaluating translations across four dimensions: accuracy, completeness, coherence, and stylistic consistency. The specific prompt used for evaluation can be found in the Appendix \ref{sec_gpt_prompt_appendix}.
The consistency between human evaluation and GPT-4o scoring can be found in the Appendix \ref{sec_gpt_human_appendix}.

\noindent \textbf{Baselines.} Our baselines include the strong general-purpose model GPT-4o, the translation-specific model Tower-7B-v0.2\cite{alves2024tower}, and the models obtained by applying SFT.

\noindent \textbf{Implementation Details.} We train our RM using Megatron framework \cite{shoeybi2019megatron} with the Qwen2.5-72B-Chat. For LLM training, we employ the Verl framework\cite{sheng2024hybridflow} with the Qwen2.5-7B-Chat. For more training details, please refer to the Appendix \ref{sec_experiment_details_appendix}.

\subsection{Main Results}
Table \ref{ASR_Result} displays the performance on the subtitle task with only the qualitative preference reward, and Table \ref{WMT_Result} presents results on WMT task with both qualitative and quantitative preference rewards.

\begin{table*}[ht]
\begin{center}
\resizebox{2.0\columnwidth}{!}{
\begin{tabular}{c|>{\columncolor{gray}}c>{\columncolor{gray}}c>{\columncolor{gray}}c>{\columncolor{gray}}c>{\columncolor{pink}}c>{\columncolor{red}}c}
\specialrule{.8pt}{0pt}{0pt}
   & \multicolumn{6}{c}{\textbf{In-Domain Subtitle \space ZH->EN}}             \\
\rowcolor{white}        & Accuracy  & Completeness & Coherence & Stylistic Consistency  & Avg           & COMETkiwi  \\ \hline
GPT-4o                  & 3.99      & 4.00         & 4.00      & 4.00                   & 4.00          & 66.82                       \\
TowerInstruct-7B-v0.2   & 2.69      & 3.69         & 3.56      & 3.71                   & 3.37          & 64.52                       \\
Qwen2.5-7B-Chat         & 2.52      & 3.4          & 3.32      &3.48                    & 3.13          & 64.03                    \\
Qwen2.5-7B-SFT          & 2.97      & 3.74         & 3.46      & 3.75                   & 3.46          & 65.89 \\ \hdashline
RIVAL-Iter0-Qual             & 2.69      & 3.4          & 3.32      & 3.48                   & 3.26          & 65.22                \\
RIVAL-Iter1-Qual             & \textbf{3.28}      & \textbf{3.86}         & \textbf{3.84}      & \textbf{3.89}                   & \textbf{3.68} & 66.27                   \\
RIVAL-Iter2-Qual             & 3.06      & 3.82         & 3.57      & 3.79                   & 3.53          & \textbf{66.49}                 \\ \hline
\specialrule{.8pt}{0pt}{0pt}
 & \multicolumn{6}{c}{\textbf{OOD Medical \space ZH->DE}}             \\
\rowcolor{white}     & Accuracy  & Completeness & Coherence & Stylistic Consistency & Avg       & COMETkiwi  \\ \hline
Qwen2.5-7B-Chat          & 2.33      & 3.19         & 3.14      & 3.28                  & 2.94      & 52.58                    \\
Qwen2.5-7B-SFT       & 2.34      & 2.99         & 2.98      & 3.08                  & 2.81      & 49.15                    \\ \hdashline
RIVAL-Iter0-Qual          & 2.48      & 3.15         & 3.10      & 3.26                  & 2.96      & 52.41                \\
RIVAL-Iter1-Qual          & 2.46      & 3.23         & 3.12      & 3.29                  & 2.99      & 53.42                   \\
RIVAL-Iter2-Qual          & 2.33      & 3.16         & 3.06      & 3.21                  & 2.93      & 51.76                 \\ \hline
\specialrule{.8pt}{0pt}{0pt}
 \end{tabular}}   
\caption{Performance comparison on in-domain translation direction (Subtitle ZH-EN) and OOD translation task (Medical ZH-DE) using GPT-4o scores and COMETKiwi. Only the qualitative preference reward is used. All results are averaged by 3 times.}
\label{ASR_Result}
\end{center}
\end{table*}

\begin{table}[ht]
\begin{center}
\resizebox{1.0\columnwidth}{!}{
\begin{tabular}{c|>{\columncolor{gray}}c>{\columncolor{gray}}c|>{\columncolor{blue}}c>{\columncolor{blue}}c}
\specialrule{.8pt}{0pt}{0pt}
In-Domain   & \multicolumn{2}{c|}{\textbf{WMT \space EN->ZH}}  & \multicolumn{2}{c}{\textbf{WMT \space ZH->EN}}    \\ \hline
\rowcolor{white}     & BLEU      & COMETkiwi      & BLEU      & COMETkiwi            \\ 
GPT-4o               & 38.98     & 75.64          & 32.4      & 73.59                 \\
TowerInstruct-7B-v0.2  & 38.69   & 71.45          & 31.61     & 71.67    \\
Qwen2.5-7B-Chat           & 31.50     & 65.41          & 25.60     & 68.61                                     \\
Qwen2.5-7B-SFT       & 38.77     & 71.39          & 32.22     & 71.88 \\ \hdashline
RIVAL-Iter0-Qual        & 33.34     & 69.64          & 28.32     & 69.91    \\
RIVAL-Iter1-Qual        & 31.74     & 70.83          & 27.65     & 71.19    \\
RIVAL-Iter2-Qual        & 30.14     & 71.91          & 26.73     & 73.28 \\ \hdashline
RIVAL-Iter0-Qual+Quant          & 34.76     & 69.12          & 29.52     & 70.16    \\
RIVAL-Iter1-Qual+Quant          & 38.62     & 69.72          & 32.90     & 72.37    \\
RIVAL-Iter2-Qual+Quant         & \textbf{39.39}     & \textbf{72.60}          & \textbf{33.42}     & \textbf{73.61}                                                                          \\ \hline
\specialrule{.8pt}{0pt}{0pt}

OOD & \multicolumn{2}{c|}{\textbf{WMT \space EN->DE}} & \multicolumn{2}{c}{\textbf{WMT \space DE->EN}}  \\ \hline
\rowcolor{white}     & BLEU      & COMETkiwi     & BLEU      & COMETkiwi                                                    \\
Qwen2.5-7B-Chat           & 27.23     & 71.53         & 36.85     & 74.18                             \\
Qwen2.5-7B-SFT       & 22.16     & 67.88         & 34.69     & 71.19                                                           \\ \hdashline
RIVAL-Iter0-Qual          & 25.15     & 71.48         & 35.17     & 73.69 \\
RIVAL-Iter1-Qual          & 24.71     & 70.93         & 34.97     & 73.06 \\
RIVAL-Iter2-Qual          & 23.96     & 69.81         & 34.82     & 72.76 \\ \hdashline
RIVAL-Iter0-Qual+Quant          & 26.44     & 71.50         & 35.35     & 73.49                                                            \\
RIVAL-Iter1-Qual+Quant          & 25.70     & 69.93         & 35.32     & 72.66                                                                \\
RIVAL-Iter2-Qual+Quant          & 25.25     & 68.74         & 35.25     & 72.23                                                                   \\ \hline
\specialrule{.8pt}{0pt}{0pt}
\end{tabular}}   
\caption{Performance comparison on in-domain WMT(EN-ZH) and OOD WMT(EN-DE) using BLEU and COMETKiwi. Both preference rewards are used. All results are averaged by 3 times.}
\label{WMT_Result}
\end{center}
\vspace{-4mm}
\end{table}

\noindent \textbf{Colloquial subtitle translation is a particularly challenging task.}
As shown in Table \ref{ASR_Result}, even TowerInstruct-7B-v0.2—a translation-specific LLM trained with approximately 20B tokens of continued pretraining—fails to perform well on this task.
According to GPT-4o evaluations, the primary issue lies in insufficient accuracy. This indicates that our ASR-based dataset contains highly diverse and substantial knowledge-intensive content, making it a relatively challenging task that has not yet been well addressed.

\noindent \textbf{As the iterations progress, the model is able to gradually discover improved translation.}
As shown in Tables \ref{ASR_Result} and Table \ref{WMT_Result}, our RIVAL method is able to progressively discover higher-quality translations through iterative optimization. 
The model outperforms current open-source translation-specific LLMs and, on certain tasks, the performance of RIVAL even exceeds that of strong general-purpose models.
These results not only demonstrate the effectiveness of our approach but also highlight the potential of iterative optimization to enable autonomous exploration and continuous adjustment toward improved translation quality.

\noindent \textbf{Different reward signals serve different purposes.}
As shown in Table \ref{WMT_Result}, our RIVAL method achieves comparable performance under both qualitative and quantitative rewards. However, when the quantitative reward is incorporated, the corresponding BLEU score improves significantly. In fact, omitting the quantitative reward leads to a drop in BLEU but an increase in COMET, suggesting that the model is able to explore effectively at the semantic level without being constrained by surface-level lexical similarity.

\noindent \textbf{In Language OOD scenarios, our approach better preserves the model's capabilities.}
As shown in Tables \ref{ASR_Result} and Table \ref{WMT_Result}, SFT severely degrades performance on OOD languages. According to GPT-based evaluation, this performance drop is primarily observed in completeness, coherence, and stylistic consistency, while accuracy remains largely unaffected. This suggests that different languages share parameters related to knowledge within the model, whereas language-specific capabilities are more tightly coupled with the language itself \cite{xu2023paradigm}.
In contrast, our RIVAL method results in significantly less degradation in OOD language performance. Moreover, we observe an interesting phenomenon: on the WMT-Medical test dataset, the performance at RIVAL-Iter1 even surpasses that of the original model. This indirectly indicates that our method encourages the exploration of more effective translation strategies, rather than relying on rote memorization of specific patterns.

In addition, we found that using only qualitative rewards can still maintain better performance than the SFT method OOD scenarios. Furthermore, on both BLEU and COMETkiwi metrics, it shows a similar pattern to the in-domain scenario: training with qualitative rewards leads to a significant decrease in character-level similarity but not at the semantic level.

\noindent \textbf{Purely the qualitative preference reward cannot guarantee a reliable optimization direction.}
As observed in Table \ref{ASR_Result} , the performance of RIVAL-Iter2 is lower than that of RIVAL-Iter1. We hypothesize that this is due to the vast exploration space inherent in open-ended generation tasks, where purely qualitative preference rewards fail to effectively constrain the model's exploration, leading to behaviors misaligned with the true reward signal. To address this, we introduce the quantitative preference reward, and as shown in Table \ref{WMT_Result}, the model’s exploration results exhibit a high degree of consistency, indicating more stable and reliable.

\subsection{Analysis}

\begin{figure*}[htbp]
\vspace{-6mm}
\centering
\subfigure[]{
\begin{minipage}[t]{0.45\linewidth}
\centering
\includegraphics[width=2.6in]{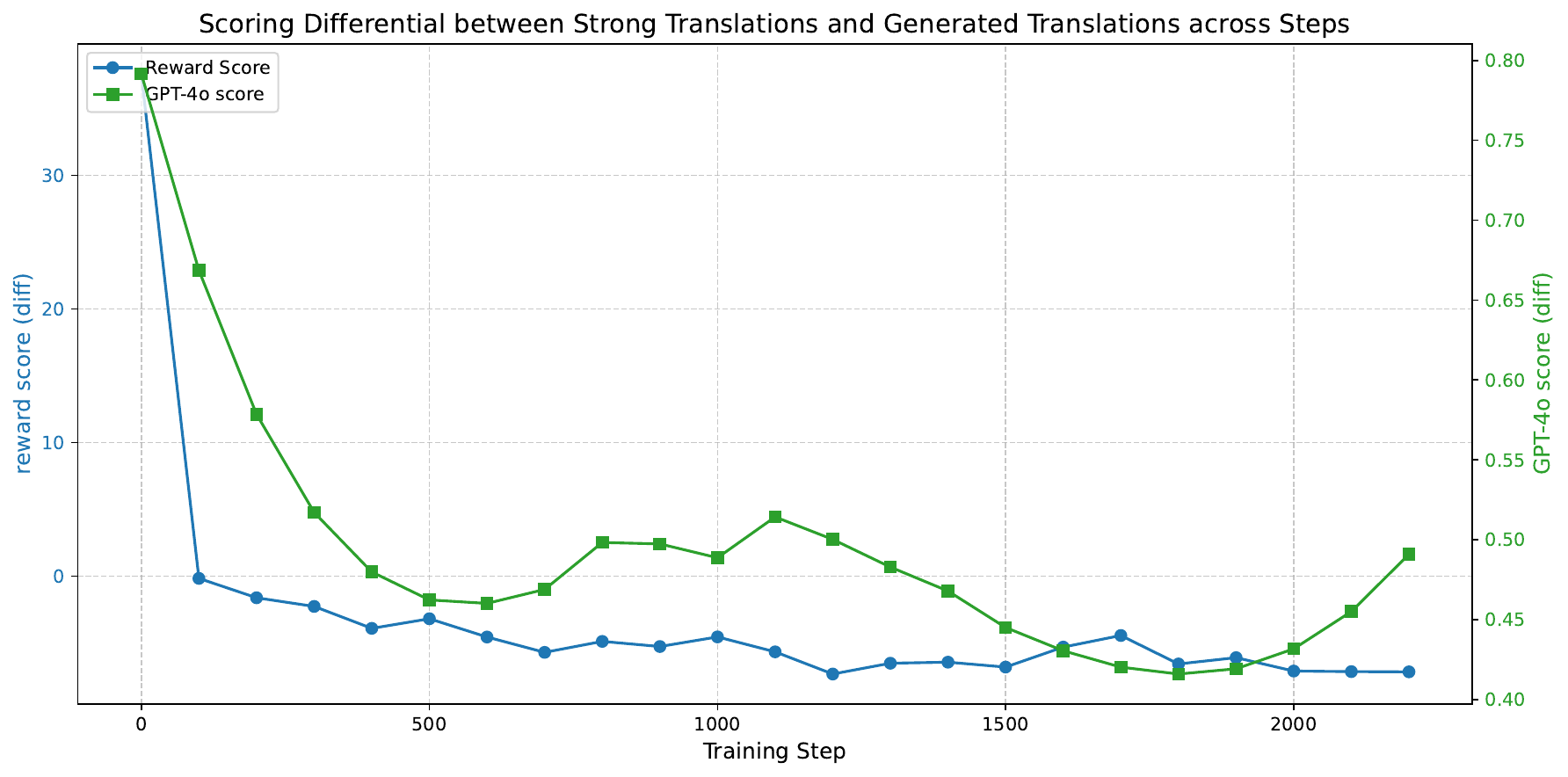}
\label{a}
\end{minipage}
}
\subfigure[]{
\begin{minipage}[t]{0.45\linewidth}
\centering
\includegraphics[width=2.6in]{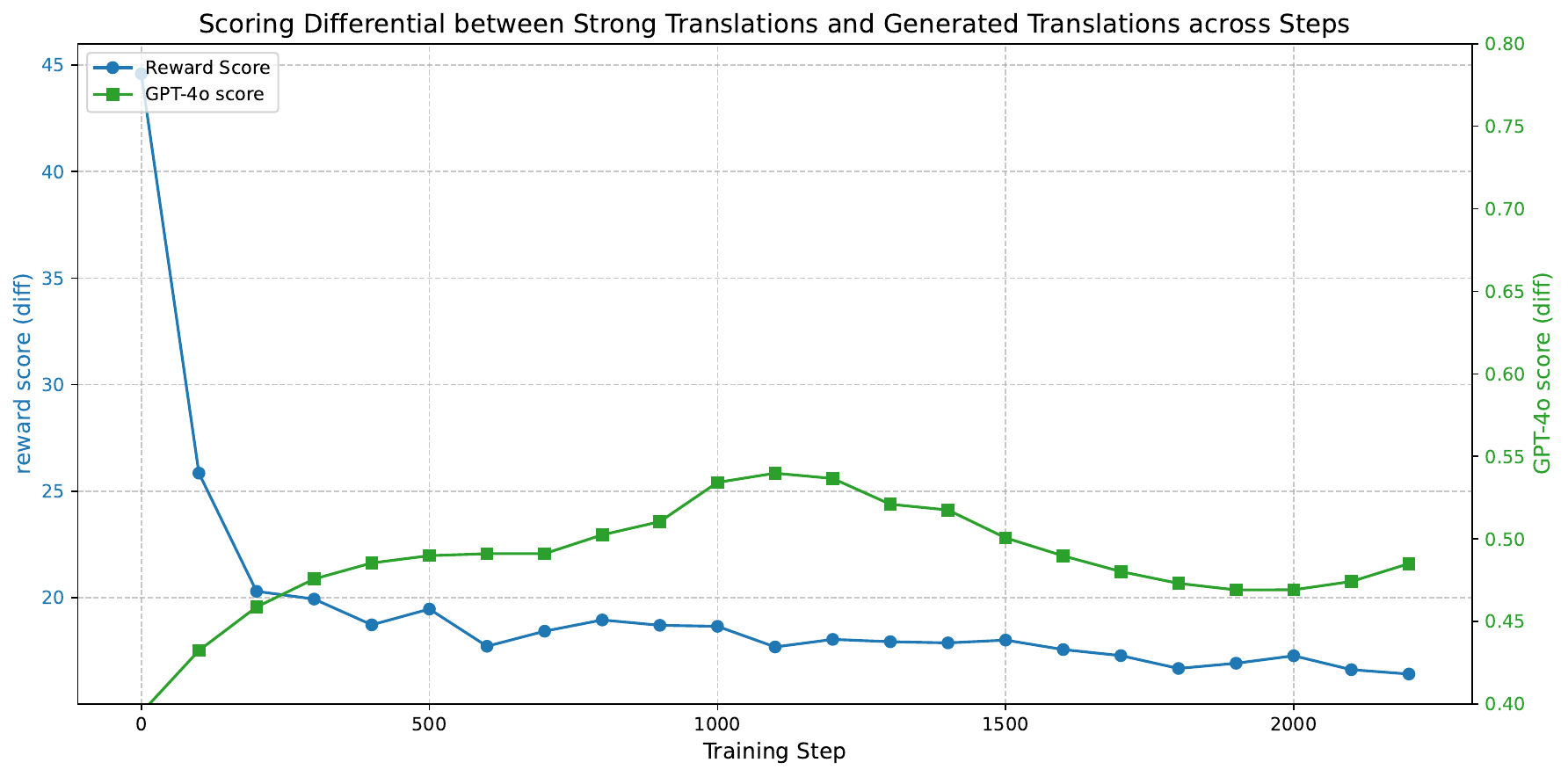}
\label{b}
\end{minipage}
}

\vspace{-2mm}
  \caption{Scoring Differential of the RM and GPT-4o between strong and weak translations.
  (a) for RIVAL-Iter1 (b) for RIVAL-Iter2.}
  
 \label{callback}
 \vspace{-2mm}
\end{figure*}

\noindent \textbf{MAE is better than MSE.}
In our experiments, we find that using MAE as the loss function for training the RM yields significantly better results than using MSE. Detailed case studies and exact figures are provided in the Appendix \ref{sec_mae_mse_appendix}. MAE reduces the error by 80\% compared to MSE and improves accuracy by 2.5\%. We attribute this phenomenon to the squared term in MSE, which greatly diminishes the loss when the target values are less than 1, making it difficult for the model to effectively learn how to fit the quantitative preference rewards. Therefore, we recommend using MAE as the loss function.

\noindent \textbf{RIVAL effectively mitigates distributional shift.}
We also plot the score differences between strong and weak translations in the last two iterations. As shown in the Figure \ref{callback}, we can observe that the score differences remain within a relatively small range throughout training and do not exhibit sudden increases in the later stages. This suggests that during effective training steps, the issue of distributional shift is significantly alleviated and iterative adversarial approach enhances the generalization capabilities of both the RM and the LLM.

\noindent \textbf{More iterations lead to better performance.}
Experimental results indicate that the model often reaches or exceeds the SFT baseline within two iterations. Continued gains observed in WMT dataset at the third iteration imply that further self-exploration may help the model surpass strong translators, which we consider a potential avenue for future work.

\section{Related works}
\label{sec_related_works_concise}

Here we present a concise overview on RL-based method for the MT(details in Appendix \ref{sec_related_works_appendix}).

\noindent \textbf{Reinforcement Learning for Machine Translation}~
Early RL studies alleviated the training–evaluation gap by directly optimizing corpus‑level BLEU and alleviating exposure bias~\cite{ranzato2016sequenceleveltrainingrecurrent, edunov-etal-2018-classical,wang-sennrich-2020-exposure}. Later work used human‑in‑the‑loop feedback for domain or user adaptation~\cite{NIPS2016_795c7a7a,nguyen-etal-2017-reinforcement}. Reward‑shaping variants inject intermediate signals, yet even token‑wise BLEU remains too coarse for fine‑grained rewards~\cite{wu-etal-2018-study,ijcai2019p331,NEURIPS2022_266c0f19}.

Modern RL‑enhanced LLMs such as OpenAI O1 and DeepSeek‑R1 achieve state‑of‑the‑art performance accross diverse benchmarks~\cite{openai2024openaio1card,deepseekR12025deepseekr1incentivizingreasoningcapability}. Motivated by these, R1‑Translator introduces chain‑of‑thought alignment for MT~\cite{he2025r1t1fullyincentivizingtranslation}; MT‑R1‑Zero adds zero‑shot hybrid rewards~\cite{feng2025mtr1zeroadvancingllmbasedmachine}. ReMedy frames MT evaluation as pairwise reward modeling~\cite{tan2025remedylearningmachinetranslation}, while severity‑weighted token rewards from XCOMET provide richer signals~\cite{ramos2025finegrainedrewardoptimizationmachine}.

\noindent \textbf{Reward Hacking in Reinforcement Learning}~
Reward hacking happens when an RL agent finds loopholes in the reward and chases the proxy instead of the real goal~\cite{weng2024rewardhack,Everitt2019RewardTP}. It has appeared in robotics, Atari‑style games, and RLHF language models, where agents loop waypoints, rewrite tests, or game metrics to boost scores without real gains~\cite{lehman2020surprising,gleave2019adversarial,denison2024sycophancy}. To prevent the above issue, recent work suggests the following methods of defence: (1) fix the reward with ensembles\cite{wortsman2022model}, adversarial training, caps, or trip‑wires~\cite{Amodei2016}; (2) harden the policy through look‑ahead planning, adversarial blinding, or indifference methods~\cite{uesato2020avoiding}; (3) add human checks, anomaly detection, and data audits after training~\cite{Pan2022TheEO,revel2025seal}.

\section{Conclusion}
\label{sec_conclusion}


Using a real-world colloquial subtitle dataset, we find that vanilla RLHF struggles to improve translation quality due to distributional shifts from offline RM. To address this, we propose RIVAL—an adversarial framework enabling iterative optimization between the RM and the LLM, incorporating both qualitative and quantitative preference rewards.
Through extensive experiments and analysis, we demonstrate the effectiveness and generalizability of this framework. 
Our work offers valuable insights for applying RL to NMT, and more broadly, serves as a reference for RL-based approaches in general post-training of LLMs.

\section*{Limitations}
In this paper, we focus on how to optimize translation performance using the RLHF paradigm, with particular emphasis on the previously underexplored task of colloquial subtitle translation.
Although our RIVAL method demonstrates superior performance over the baselines in experiments, we have not yet explored its full potential. We believe that with additional rounds of iteration, the model could achieve performance comparable to, or even surpass, that of strong translators.
Moreover, we believe that this iterative adversarial optimization approach can be extended to more general post-training settings. Given its generalizability, we hope it can serve as an effective alternative to supervised fine-tuning.



\bibliography{anthology, custom}
\bibliographystyle{acl_natbib}

\appendix
\section{Subtitle Case}
\label{sec_subtitle_appendix}
\colorbox{gray!80}{
\begin{minipage}{0.45\textwidth}
\footnotesize
\begin{CJK}{UTF8}{gbsn}
\noindent \textbf{Source}: 你将一个视频中的多个语音识别文本逐条翻译成英文。输入为一个json格式，key为序号，value为待翻译的语音文本，一共有10个文本。示例如下: \\ 输入: \{"1": "xxx", "2": "xxx"\}\\ \\ 输出: \{"1": "xxx", "2": "xxx"\}\\ \\现在请根据上述要求完成如下片段的翻译，输出一共10个翻译后的结果，需要和输入一一对应。直接输出翻译后的英文，不要进行任何解释。输入: \{"1": "牛老师惊恐的说：你大的要来了？我点点头：要来了，我瞒不住了。", "2": "其实我是一名鬼杀队队员。牛老师大怒：你是不是有病？我都这样了，你还不放过我？", "3": "而且还要 COS 鬼灭之刃。我叹了口气：You KNOW NOTHING ，牛牛桑，鬼灭之刃其实不是虚构的，而是一部纪录片。", "4": "世界上是真的有鬼，也是真的有鬼杀队员。", "5": "如果今天不是我及时出现，大爷们就要白发人送黑发人了。", "6": "刘老师说：那问题是我刚染了黄毛呀，不是黑发人呀。","7": "我说：我说的是你的脸，印堂发黑去非洲怎么着也得当一个王妃。", "8": "之所以你认为这个世界上没有鬼存在，都是我们这些鬼杀队队员的功劳，把你这些普通人类保护的太好。", "9": "当然，也正因为我们跟鬼残酷的战斗了几百年，以至于人才凋落，很多鬼都管不过来，才让你被鬼白嫖。", "10": "牛老师说：你快拉倒吧，你这人全身上下只有嘴硬，就你还杀鬼，你只会鬼吹灯。"\}\\输出: 
\end{CJK}
\end{minipage}
}
\colorbox{red!40}{
\begin{minipage}{0.45\textwidth}
\noindent \textbf{Strong Translation}: \{"1": "Teacher Niu said in horror: Your big one is coming? I nodded: It's coming, I can't hide it anymore.", "2": "Actually, I am a member of the Demon Slayer Corps. Teacher Niu was furious: Are you sick? I'm already like this, and you still won't let me go?", "3": "And I have to COS Demon Slayer. I sighed: You KNOW NOTHING, Niu Niu-san, Demon Slayer is not fictional, it's actually a documentary.", "4": "There really are demons in the world, and there really are demon slayers.", "5": "If I hadn't appeared in time today, the old men would have had to send off the young.", "6": "Teacher Liu said: But the problem is I just dyed my hair yellow, I'm not a young person with black hair.", "7": "I said: I'm talking about your face, with a darkened forehead, you could at least be a princess in Africa.", "8": "The reason you think there are no demons in this world is all thanks to us demon slayers, protecting you ordinary humans too well.", 
\end{minipage}
}
\colorbox{red!40}{
\begin{minipage}{0.45\textwidth}
"9": "Of course, it's also because we've been fighting brutally with demons for hundreds of years, leading to a decline in talent, and many demons are left unchecked, which is why you got taken advantage of by demons.", "10": "Teacher Niu said: Stop it, you're all talk and no action, you killing demons? You only know how to blow hot air."\}"
\end{minipage}
}
\colorbox{gray!80}{
\begin{minipage}{0.45\textwidth}
\footnotesize
\begin{CJK}{UTF8}{gbsn}
\noindent \textbf{Source}: "请你将一个视频中的多个语音识别文本逐条翻译成英文。输入为一个json格式，key为序号，value为待翻译的语音文本，一共有10个文本。示例如下:\\ 输入: \{"1": "xxx", "2": "xxx"\}\\输出: \{"1": "xxx", "2": "xxx"\}\\现在请根据上述要求完成如下片段的翻译，输出一共10个翻译后的结果，需要和输入一一对应。直接输出翻译后的英文，不要进行任何解释。\\ 输入: \{"1": "啊哈。但是你听到后面跟的是个降调的感觉。听啊，", "2": "啊，再来一下。", "3": "当然了，倒水最主要声音最响就是这个叭叭这种声音。", "4": "贯穿始终有一种大珠小珠落玉盘的感觉，它呢，其实就是这种入水的声音。就是这，", "5": "是吧？", "6": "我应该找 B .Boxer 酋长来录这一期的。", "7": "总之呢，倒一杯水有三种声音诶，嘚嘚吧嘚吧。", "8": "但是在科学上，你不能用得儿劲儿蹦来描述谁的声音。听着不太聪明的样子。", "9": "声音的本质是物体的振动。那这三种声音都是啥在振动呢？那个声调的怎？", "10": "N 是杯子里面水面上方空气柱的振动。也就是我们小学二年级学过的海姆霍兹共振。"\}\\输出: " 
\end{CJK}
\end{minipage}
}
\colorbox{red!40}{
\begin{minipage}{0.45\textwidth}
\noindent \textbf{Strong Translation}: "\{"1": "Aha. But you hear that the following part has a falling tone. Listen,", "2": "Ah, let's do it again.", "3": "Of course, the loudest sound when pouring water is this kind of 'ba ba' sound.", "4": "Throughout, there is a feeling of big and small beads falling on a jade plate. It is actually the sound of water entering. It's this,", "5": "Right?", "6": "I should have invited B.Boxer Chief to record this episode.", "7": "In short, there are three kinds of sounds when pouring a glass of water, de de ba de ba.", "8": "But scientifically, you can't describe someone's sound as 'de er jin er beng'. It doesn't sound very smart.\", "9": \"The essence of sound is the vibration of objects. So what are these three sounds vibrating? What about the pitch?", "10": "N is the vibration of the air column above the water surface inside the cup. This is the Helmholtz resonance we learned in second grade."\}"
\end{minipage}
}

\section{GPT-4o Evaluation Prompt} 
\label{sec_gpt_prompt_appendix}

\colorbox{gray!20}{
\begin{minipage}{0.45\textwidth}
\footnotesize
\begin{CJK}{UTF8}{gbsn}
\textbf{GPT-4o Evaluation Prompt 如下：}

请你以公正的评判者的身份，结合参考译文，对翻译文本的效果进行评估。请保持一致的评估标准。你需要按照下面的维度对翻译文本进行评估，根据每个维度的要求描述，给每个维度打出一个1\~4的分数。

\textbf{维度1：准确性：词句+专有名词翻译是否准确}
\begin{itemize}
\item 1分：翻译严重偏离原义，3处以上词语存在事实性错误或语义矛盾。
\item 2分：翻译部分偏离原义，2-3处词语存在事实性错误或语义矛盾。
\item 3分：1处词语存在事实性错误或语义矛盾，或是无词语翻译错误，仅有语义错误。
\item 4分：译文完全准确，无事实性错误，且不影响用户感知。
\end{itemize}

\textbf{补充说明：}
1. 如果翻译存在漏翻，如果未导致句子主要含义发生变化，则只扣完整性，不扣准确性分数；
2. 原文的错别字、断句不需要结合上下文进行纠正，直接按照原文文本翻译即可，翻译正确即不扣准确性分数。
3. 增添部分逻辑合理，不做准确性扣分。
\textbf{维度2：完整性：是否漏翻关键词句}
\begin{itemize}
\item 1分：句子主要信息和逻辑关系严重缺失，主从句不完整，数词、动词及动词受者大量缺失，达3处以上遗漏问题，严重影响原义理解。
\item 2分：句子主要信息部分缺失主句和从句（非状语、非补语）部分未保留主谓语，关键成分（数词、动词及其受者）未完整保留，达2-3处遗漏问题，不影响整体理解。
\item 3分：句子主要信息部分缺失，主句和从句（非状语、非补语）部分未保留主谓语，关键成分（数词、动词及其受者）仅小部分遗漏，仅1处遗漏问题，不影响整体理解。
\item 4分：内容完整性：确保句子的主要信息和逻辑关系100\%完整。主从句完整性：主句和从句（非状语、非补语）必须保留主谓语。仅允许进行浓缩或代词替换。其他成分处理：宾语、补语、插入语和修饰语可以进行压缩性意译或删除。关键成分保留：数词、动词及其受者不可省略，必须完整保留。
\end{itemize}
\textbf{补充说明：}
1）请严格对比原文，看主要信息是否均翻译完成，可以接受对语义重复的修饰语做删除，不扣完整性分数；
2）原文的错别字、断句错误导致的问题，不扣完整性分数。
\textbf{维度3：连贯性：上下文衔接是否连贯}
\begin{itemize}
\item 1分：语法错误频发，上下文转折生硬，逻辑断裂，严重影响用户理解。
\item 2分：多处语序混乱或句式杂糅，上下文衔接略生硬，关联词使用不准确，需反复理解
\item 3分：整体连贯，仅存在部分的生硬衔接，存在过长修饰语或出现少量语句重复。
\item 4分：语句衔接自然，表达顺畅，符合口语表达习惯。
\end{itemize}
\end{CJK}
\end{minipage}
}

\colorbox{gray!20}{
\begin{minipage}{0.45\textwidth}
\footnotesize 
\begin{CJK}{UTF8}{gbsn}
\textbf{补充说明：}
1）连词、独立语、过渡性语句等的省略，如果对句子内容产生影响，则只扣完整性分数，连贯性不扣分；
2）因为原文错别字、断句导致的问题，不对此维度扣分。
\textbf{维度4：风格一致性：上下文叙述风格是否一致}
\begin{itemize}
\item 1分：翻译内容与视频语境不符，3处以上语法结构、用词与视频风格不一致，使用过于突兀。
\item 2分：翻译内容在书面语和口语之间切换，2-3处用词和句法结构不符合语境。
\item 3分：上下文风格大体一致，自然流畅，仅1处用词和句法结构不符合语境。
\item 4分：上下文内容呈现出统一的口语或书面语的特征，整体内容风格明显。
\end{itemize}

\textbf{补充说明：}
首先判断视频文本的场景为书面还是口语场景，后判断语法结构、用词是否符合场景要求。

我们会提供原文、参考译文、待评估的翻译文本，你需要依照前面的每个维度标准，分别给出翻译文本在各个维度的打分并解释理由。请注意，评分应尽可能严格。如果样例在某个维度上同时符合低分和高分的评价标准，应优先选择低分，而非较高的分数。只有当样例完全符合高分标准时，才能给予高分。你的回答以 JSON 格式输出结果，严格按照如下格式返回：

\{"维度1": \{"解释": "解释", "打分": 1\}\}

【以下是原文】
\{origin\_text\}

【以下是参考译文】
\{reference\_text\}

【以下是待评估的翻译文本】
\{translate\_text\}
\end{CJK}
\end{minipage}
}

\section{GPT4o vs Human}

\label{sec_gpt_human_appendix}
\begin{figure}[h]
\centering
  \includegraphics[width=2in]{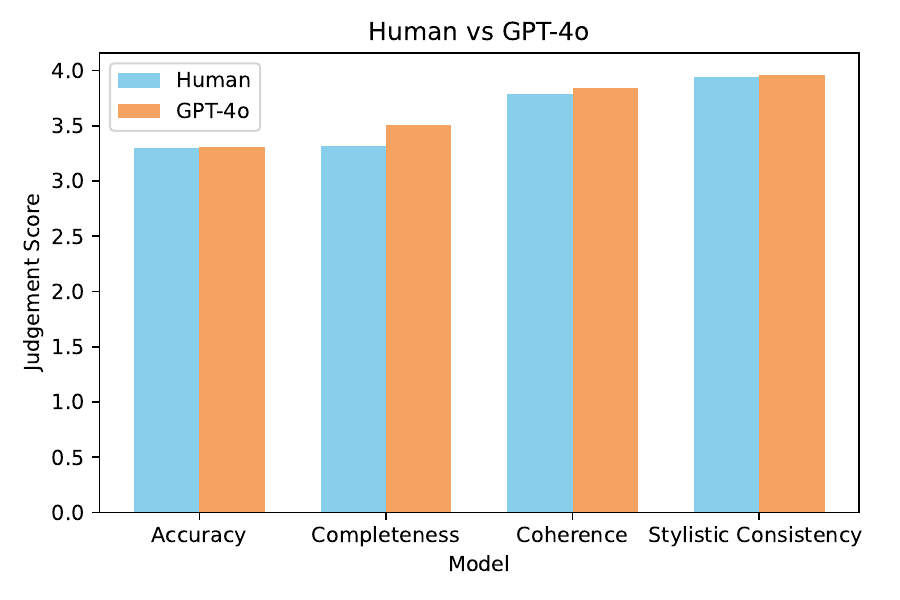}
  \caption{Human Evaluation vs GPT-4o Evaluation}
 \label{HumanVSGPT4o}
\end{figure}

To assess the reliability of GPT-4o in evaluating translation quality, we conduct a comparative analysis between human and model-based judgments on the subtitle dataset. Specifically, we invite three human annotators—none of whom are involved in this study—to independently rate the translation outputs. The three volunteers have received a solid higher education, and we provided them with detailed instructions on the annotation task, including scoring criteria and evaluation standards.
The average scores assigned by these human evaluators are then compared to those produced by GPT-4o. As presented in Figure~\ref{HumanVSGPT4o}, GPT-4o's assessments show a high degree of consistency with human judgments, suggesting that it can serve as a reliable proxy for human evaluation in this context. Given the strong alignment and the practicality of automated evaluation at scale, we adopt GPT-4o as the primary evaluator for all translation quality assessments conducted in this work.

\section{Implementation Details}
\label{sec_experiment_details_appendix}
Below are some specific details of our model training.

Our RM training framework is built on Megatron. We use the Qwen2.5-72B-Chat model as the initialization. The training is conducted with a batch size of 256, using a cosine learning rate scheduler with an initial learning rate of 5e-6. All models are trained on 64  Huawei's Ascend 910B NPUs.

Our RL training framework is based on the Verl framework. We use the Qwen2.5-7B-Chat model as the initialization for RL training. During training, we configure a batch size of 16 and perform 16 rollouts per prompt using the GRPO algorithm. The learning rate is initialized at 1e-8, and a cosine scheduler with warm-up is applied toward the final iteration. Sampling is conducted with a temperature of 1.0, and the maximum generation length is limited to 1,024 tokens. The KL penalty coefficient $\beta$ is set to 0, effectively removing the KL constraint relative to the reference policy. 
The PPO clipping range $\epsilon$ is fixed at 0.2. All models are trained for one epoch using 8 NVIDIA H800 80G GPUs.

For the subtitle task, the RM training data consists of 60,000 samples, and the RL training data contains 60,000 samples.
For the WMT task, the RM training data includes about 50,000 samples, and the RL training data comprises about 30,000 samples.

\section{MAE vs MSE case}
\label{sec_mae_mse_appendix}
\begin{CJK}{UTF8}{gbsn}
\textbf{Source}: 请将以下内容翻译为英文: 他强调,遵照秘鲁宪法和国际人权文件,秘鲁政府不会强迫个人接受生殖健康和计划生育方案的服务。\\
\end{CJK}

\noindent  \textbf{Strong Translation}: He emphasized that in keeping with the country’s Constitution and with the international documents on human rights, the Government of Peru did not coerce individuals who availed themselves of the services of the reproductive health and family planning programme.\\
\noindent \textbf{Weak Translation}: He emphasized that, in accordance with Peruvian constitution and international human rights documents, the Peruvian government would not force individuals to accept services related to reproductive health and family planning. \\
$\text{strong bleu score} = 1$ \\
$\text{weak bleu score} = 0.6404765601431773$ \\
$\text{mae strong bleu score} = 0.9680025577545166$ \\
$\text{mae weak bleu score} = 0.6667302250862122$ \\
$\text{mse strong bleu score} = 1.941943645477295$ \\
$\text{mse weak bleu score} = 1.9110959768295288$ \\

\begin{CJK}{UTF8}{gbsn}
\noindent \textbf{Source}: 请将以下内容翻译为中文: 235. At the first regular session of 1999 of the Administrative Committee on Coordination, in April, the members concluded that, to meet the challenges of globalization, the United Nations system needed to cooperate more effectively with the private sector and civil society, as well as with Governments.\\
\noindent \textbf{Strong Translation}: 235. 行政协调会成员在1999年4月第一届常会上总结指出,为应付全球化的挑战,联合国系统必须更有效地与私营部门、民间社会和各国政府合作。\\
\noindent \textbf{Weak Translation}: 在1999年4月召开的协调委员会第一次定期会议上，成员们得出结论认为，为了应对全球化带来的挑战，联合国系统需要更加有效地与私营部门、民间社会以及各国政府进行合作。 \\
\end{CJK}
$\text{strong bleu score} = 1$ \\
$\text{weak bleu score} = 0.10885796200376416$ \\
$\text{mae strong bleu score} = 1.09672212600708$ \\
$\text{mae weak bleu score} = 0.2919744849205017$ \\
$\text{mse strong bleu score} = 1.3286265134811401$ \\
$\text{mse weak bleu score} = 1.2083359956741333$ \\

\begin{table}[h]
\centering
\begin{tabular}{c|cc}
\specialrule{.8pt}{0pt}{0pt}
            & Quantitative Error      & Qualitative Acc      \\ \hline 
   MAE      & 0.19                    & 99.5                    \\
   MSE      & 0.93                    & 97.0                       \\
\specialrule{.8pt}{0pt}{0pt}
\end{tabular}
\caption{Comparison Between MAE and MSE for Training the RM.}
\label{MAEvsMSE}
\vspace{-2mm}
\end{table}

\section{Related works}
\label{sec_related_works_appendix}

\textbf{Reinforcement Learning for Machine Translation}~Early works employed RL to bridge the gap between training on token‑level log‑likelihoods and evaluating on corpus‑level, non‑differentiable metrics like BLEU~\cite{ranzato2016sequenceleveltrainingrecurrent,edunov-etal-2018-classical}, while also mitigating exposure bias in autoregressive decoders~\cite{wang-sennrich-2020-exposure}. Subsequent research repurposed RL for domain or user adaptation by replacing engineered objectives with human feedback in human‑in‑the‑loop workflows~\cite{NIPS2016_795c7a7a,nguyen-etal-2017-reinforcement}. Recent methods add reward shaping~\cite{wu-etal-2018-study,ijcai2019p331,NEURIPS2022_266c0f19}, injecting intermediate signals alongside final‑step BLEU, yet partial or token‑wise BLEU still fails to capture subtle semantic and contextual differences, making it a poor choice for fine‑grained reward design. 

Nowadays, ground‑breaking RL-based LLMs typified by OpenAI’s O1~\cite{openai2024openaio1card} and DeepSeek‑R1~\cite{deepseekR12025deepseekr1incentivizingreasoningcapability} deliver state‑of‑the‑art results on diverse benchmarks, drawing considerable attention from the research community.  Extending this progress, R1‑Translator~\cite{he2025r1t1fullyincentivizingtranslation} is the first to incorporate human‑aligned chain‑of‑thought reasoning into general machine translation through RL. MT‑R1‑Zero~\cite{feng2025mtr1zeroadvancingllmbasedmachine} pushes the paradigm further by introducing zero‑shot RL with a hybrid rule‑and‑metric reward for translation tasks. Meanwhile, ReMedy~\cite{tan2025remedylearningmachinetranslation} reconceptualizes machine‑translation evaluation as a reward‑modeling problem, training on pairwise preferences to avoid the noise of absolute quality ratings. Complementary work~\cite{ramos2025finegrainedrewardoptimizationmachine} replaces sparse sentence‑level feedback with severity‑weighted, token‑level rewards computed by XCOMET, yielding a more informative learning signal.

\textbf{Reward Hacking in Reinforcement Learning}~Reward hacking~\cite{weng2024rewardhack, Everitt2017ReinforcementLW, Everitt2019RewardTP, Langosco2021GoalMI, Pan2022TheEO} occurs when a reinforcement‑learning (RL) agent exploits flaws in its reward signal and optimizes the proxy rather than the true task goal. This vulnerability stems from the practical difficulty of defining a complete, noise‑free reward in complex or partially observed environments. Documented cases span robotics~\cite{lehman2020surprising, christiano2017deep}, Atari‑style games~\cite{Bansal2017EmergentCV, gleave2019adversarial}, and RLHF pipelines for language models~\cite{Paulus2017ADR, denison2024sycophancy}, with agents looping around waypoints, rewriting unit tests, or gaming automatic metrics to inflate scores without real progress. Recent research therefore frames reward hacking as a safety and alignment challenge and proposes layered defences: (i) reward‑side fixes—adversarial or ensemble reward models, potential‑based shaping, reward capping, and “trip‑wire” signals that trigger intervention~\cite{Amodei2016}; (ii) policy‑side safeguards—model look‑ahead, adversarial blinding, or indifference techniques to block exploits~\cite{uesato2020avoiding}; and (iii) post‑hoc monitoring—decoupled human approval, anomaly detection on trajectories, and systematic data audits to surface misalignment early~\cite{Pan2022TheEO, revel2025seal}. Together, these strategies emphasise adaptive, multi‑stage protection rather than any single remedy.

\end{document}